\documentclass[runningheads]{llncs}
\usepackage[T1]{fontenc}
\usepackage{graphicx}
\usepackage{amsmath}
\usepackage{amssymb}
\usepackage{marvosym}

\begin{document}

\title{An improved EfficientNetV2 for garbage classification}
\author{Wenxuan Qiu\inst{1} \and
Chengxin Xie\inst{1} \and
Jingui Huang\inst{1,\textsuperscript{\Letter}}}
\authorrunning{W. Qiu et al.}
\institute{College of Information Science and Engineering,Hunan Normal University, Changsha, 410081, Hunan,China\\
\email{hjg@hunnu.edu.cn} 
}
\maketitle              
\begin{abstract}
This paper presents an enhanced waste classification framework based on EfficientNetV2 to address challenges in data acquisition cost, generalization, and real-time performance. We propose a Channel-Efficient Attention (CE-Attention) module that mitigates feature loss during global pooling without introducing dimensional scaling, effectively enhancing critical feature extraction. Additionally, a lightweight multi-scale spatial feature extraction module (SAFM) is developed by integrating depthwise separable convolutions, significantly reducing model complexity. Comprehensive data augmentation strategies are further employed to improve generalization. Experiments on the Huawei Cloud waste classification dataset demonstrate that our method achieves a classification accuracy of 95.4\%, surpassing the baseline by 3.2\% and outperforming mainstream models. The results validate the effectiveness of our approach in balancing accuracy and efficiency for practical waste classification scenarios.

\end{abstract}
\section{Introduction}
Waste classification stands as a cornerstone technology for advancing urban sustainability and environmental protection. With the accelerating pace of global urbanization, waste production continues to rise~\cite{2024A}, making waste management a worldwide challenge. Conventional disposal methods, such as landfilling and incineration, not only consume vast land resources but also cause severe environmental pollution, including soil degradation, water contamination, and air pollution~\cite{2023A}. Traditional reliance on manual sorting is inefficient and costly, failing to meet the demands of modern smart cities for resource recovery and ecological preservation. There is an urgent societal need for effective waste management solutions to enhance resource recycling rates, mitigate environmental harm, and promote sustainable development.

In recent years, rapid advancements in artificial intelligence (AI) have introduced novel approaches to waste classification~\cite{2019Fully}. Computer vision-based automated classification technologies, capable of efficiently processing massive volumes of waste images, significantly improve accuracy, speed, and operational independence, thereby providing robust technical support for waste resource utilization. Notably, AI-driven solutions have also been successfully applied in domains such as person generation~\cite{shen2024imagpose}, video synthesis~\cite{shen2025long}, and multimodal retrieval~\cite{shen2023pbsl}, offering useful insights for enhancing feature representation and attention modeling in waste classification tasks.

Early studies primarily relied on manually designed feature extraction methods. These methods rely on manually extracting image features such as color, texture, or shape (e.g., color histograms, SIFT~\cite{2004Distinctive}, HOG~\cite{0Body} shape features) and subsequently employ traditional classifiers (e.g., Support Vector Machines (SVM)~\cite{1995Support}, K-Nearest Neighbors (KNN)~\cite{2001A}) for category determination. While stable in specific scenarios, these methods exhibited inherent limitations due to their dependence on human priors: handcrafted features were sensitive to complex backgrounds, lighting variations, resulting in weak generalization; and limited feature representation hindered the capture of fine-grained semantic information, leading to classification accuracy bottlenecks.

In recent years, end-to-end deep learning methods have emerged as the mainstream. These approaches leverage convolutional neural networks (CNNs) to autonomously learn high-level semantic features, substantially improving classification performance. For instance, ResNet~\cite{2016Deep} addressed deep network training challenges through residual structures and has been widely adopted for waste image classification. A study based on ResNet50~\cite{2021Research} designed a lightweight waste classification model that skillfully integrates depthwise and group convolutions to reduce computational costs and parameter counts while improving accuracy through channel attention mechanisms. Lightweight architectures like MobileNet~\cite{2017MobileNets} and EfficientNet~\cite{2019EfficientNet} further optimized computational efficiency, facilitating deployment on edge devices. However, existing methods still face two major challenges: high computational costs of complex models limit their applicability in resource-constrained scenarios; and insufficient robustness to background noise, small objects, and class-imbalanced data leads to performance degradation in real-world settings. Although some studies have attempted improvements via attention mechanisms (e.g., SE modules) or data augmentation strategies, the balance between feature extraction efficiency and model lightweighting remains unresolved.

To address these issues, this article proposes an enhanced EfficientNetV2 model that integrates a novel attention mechanism, optimized module structures, and data augmentation strategies to improve classification accuracy while maintaining computational efficiency. The experimental results demonstrate the superiority of our method over existing baseline models, offering an advanced solution for the classification of waste in complex scenarios. The key contributions are as follows:
\begin{itemize}
\item[$\bullet$] Waste classification facilities often operate under limited computational resources. To balance accuracy and efficiency, we propose a novel CE channel attention mechanism module. This design addresses the shortcomings of the original SE module in EfficientNetV2, which suffers from incomplete feature extraction and excessive complexity. 
\item[$\bullet$] Existing models predominantly focus on single or few targets with simple backgrounds, struggling to adapt to real-world scenarios characterized by diverse waste types and cluttered environments. We enhance the multiscale spatial feature extraction module (SAFM) to enable richer and more detailed feature capture, thereby improving accuracy and generalization.
\item[$\bullet$] Publicly available waste classification datasets are scarce and image quality is often degraded by factors such as lighting variations and object deformation. To address this, we employ rotation, translation, noise injection, and other data augmentation techniques to enhance data diversity, significantly boosting model robustness and generalization capabilities.
\end{itemize}

\section{Related Work}

\subsection{EfficientNetV2 and Its Limitations}

EfficientNetV2~\cite{2021EfficientNetV2} is a widely adopted convolutional neural network architecture that achieves a balance between accuracy and computational efficiency through compound scaling strategies and progressive training. Its backbone incorporates the MBConv and Fused-MBConv modules, leveraging depthwise separable convolutions and SE-Net channel attention to enable compact and high-performing models. Compared with its predecessor, EfficientNetV2 demonstrates faster training convergence and improved inference efficiency, particularly on large-scale datasets.
However, EfficientNetV2 still exhibits limitations when addressing complex scenes or multi-scale inputs. In particular, the channel attention mechanism employed by SE-Net relies on global average pooling followed by two fully connected layers, which may weaken fine-grained feature distinctions and introduce additional computational overhead~\cite{2020ECA}. Furthermore, its global context modeling capability is constrained, limiting its effectiveness in tasks requiring spatial adaptability or dynamic modulation.

To address these constraints, recent works in remote sensing and aerial object detection have introduced efficient attention modules and feature pyramid optimizations. For example, the Selective Frequency Interaction Network (SFINet)~\cite{weng2024enhancing} improves aerial object detection by dynamically adjusting frequency components of features, while LR-FPN~\cite{li2024lr} enhances spatial localization through position-aware pyramid refinements. These insights inspire the need for adaptive and lightweight attention designs for waste classification tasks with complex spatial contexts.

\subsection{Spatially-Adaptive Feature Modulation (SAFM)}

Spatially-Adaptive Feature Modulation (SAFM)~\cite{sun2023spatially} is a convolution-based module that mimics the multi-head attention mechanism of Vision Transformers (ViTs) while maintaining high efficiency. It consists of a multi-scale feature generator and a dynamic spatial attention unit, which collectively improve the network’s capacity to capture contextual details across scales and spatial positions. Compared with standard self-attention, SAFM is more computationally friendly and suitable for deployment on edge devices.
Recent research across vision-language tasks and cross-modal generation has underscored the effectiveness of spatially adaptive modules. For example, Shen et al.~\cite{shen2024boosting} demonstrate improved consistency in story visualization through contextual modulation, while IMAGDressing~\cite{shen2024imagdressing} integrates controllable spatial attention for personalized virtual try-on. Similarly, Gao et al.~\cite{gao2024exploring} employ adaptive latent warping to refine pose-conditioned virtual dressing, showcasing the value of spatial adaptivity in fine-grained recognition and generation.

In this study, we integrate an optimized SAFM module into EfficientNetV2 to enhance its performance on waste classification. Specifically, SAFM is placed after each Fused-MBConv block to reinforce hierarchical spatial modulation. To maintain lightweight properties, standard convolutions in SAFM are replaced with depthwise separable convolutions, enabling fine-grained spatial awareness while preserving inference efficiency. The design is inspired by recent advancements in pose-guided generation~\cite{shen2023advancing} and multimodal image synthesis~\cite{shen2024boosting}, where adaptive spatial feature handling significantly improves visual precision and robustness.

\begin{figure}
\centering
\includegraphics[width=\linewidth]{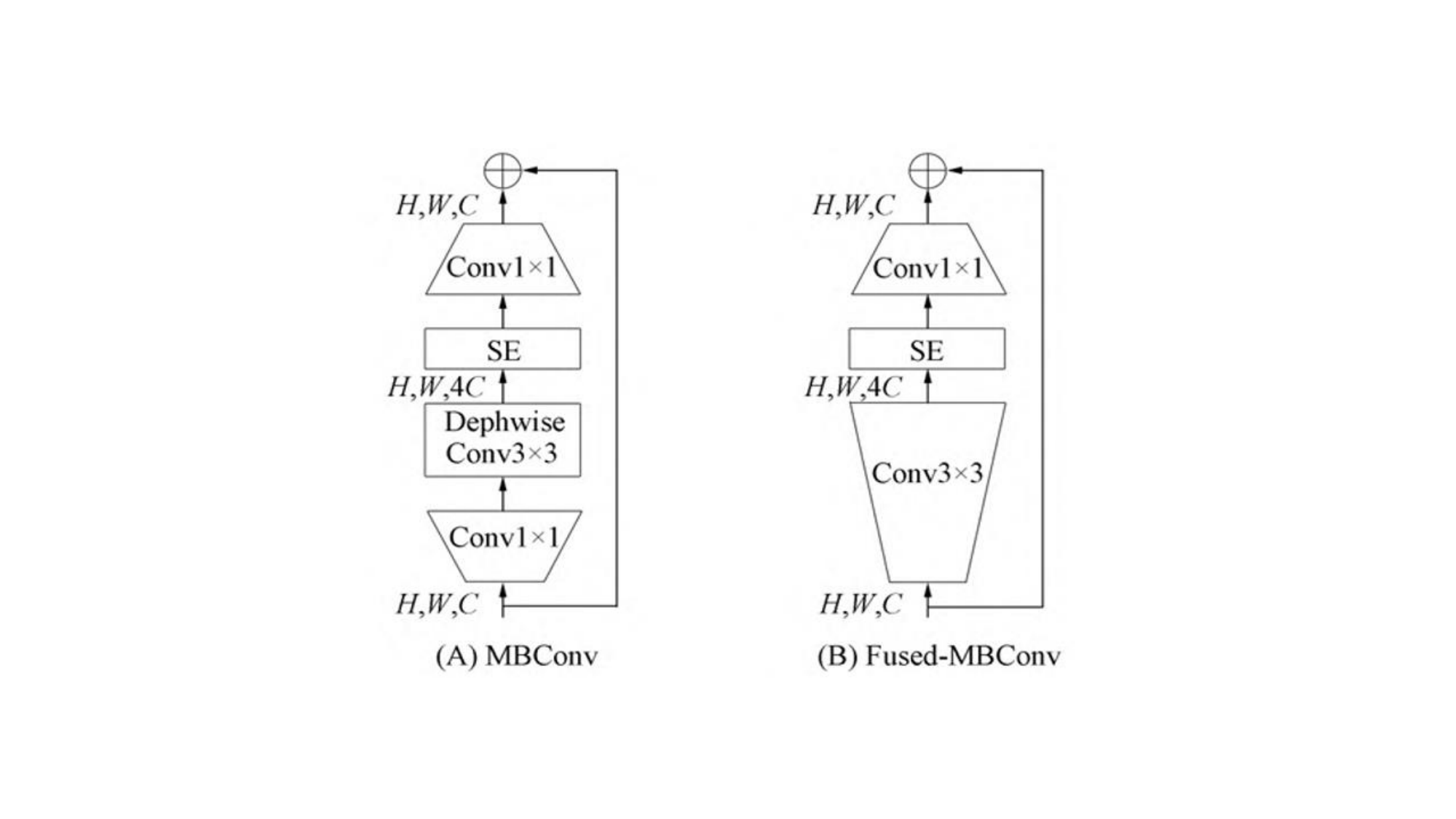}
\caption{Structure of the MBConv and Fused-MBConv module in EfficientNetV2} \label{fig1}
\end{figure}

\section{Proposed Method}
\subsection{Model}
\begin{figure}
\centering
\includegraphics[width=\linewidth]{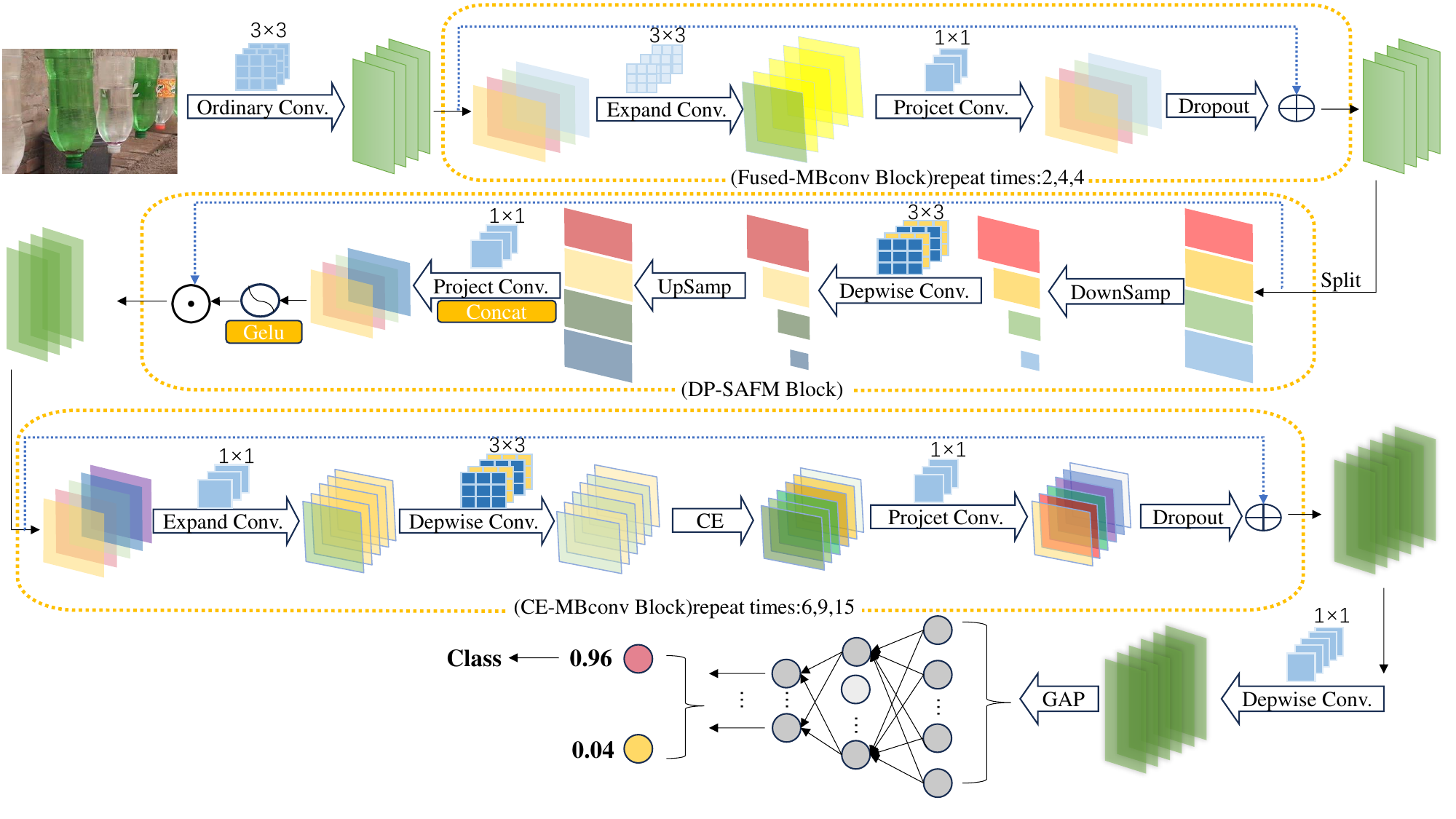}
\caption{Structure of the CE-EfficientNetV2 Model} \label{fig2}
\end{figure}

Building upon the EfficientNetV2 backbone network, our proposed architecture incorporates two key enhancements. First, considering that deeper network layers encode more complex and abstract features, we replace the SE attention mechanism in MBConv blocks with the Channel Efficient(CE) attention module to improve semantic feature capture. Second, we insert the SAFM module after Fused-MBConv layers. The multi-scale feature extraction and dynamic modulation mechanisms of SAFM strengthen the feature representation capability of EfficientNetV2, ensuring discriminative and robust feature generation. The overall Channel Efficient EfficientNetV2(CE-EfficientNetV2) architecture is illustrated in Figure \ref{fig2}.
\subsection{Channel Efficient(CE) attention module}
To address the limitations of SE-Net, we propose a novel attention mechanism that enhances local feature extraction while reducing parameter redundancy through multi-scale pooling operations and a lightweight channel mixing strategy. The detailed design is as follows:\\
\indent To overcome the limitations of global average pooling, we introduce multi-scale pooling by concurrently applying global average pooling and global max pooling to capture different feature statistics.\\
\indent For an input feature map $F\in \bbbr^{H\times W \times C }$,global average pooling and global max pooling operations are first applied to each channel to compute the average and maximum feature values per channel. This yields two channel-wise vectors, $F_{avg}\in \bbbr^{1\times 1 \times C }$ and $F_{max}\in \bbbr^{1\times 1 \times C }$ representing the global average and maximum features of each channel, respectively.

\begin{align}
    F_{avg}=AvgPool(F),\\
    F_{max}=MaxPool(F).
\end{align}

\indent MaxPool emphasizes salient local features, while AvgPool retains holistic spatial information. The element-wise summation of these pooled results generates enriched feature representations, enhancing the model’s sensitivity to fine-grained details.\\
\indent Lightweight Channel Mixing:To mitigate parameter redundancy from fully connected layers, we design a lightweight channel mixing strategy. A multi-layer perceptron (MLP), structured as Conv-ReLU-Conv layers, learns channel dependencies from $F_{avg}$ and $F_{max}$, producing refined vectors $A_{avg}\in \bbbr^{1\times 1 \times C }$ and $A_{max}\in \bbbr^{1\times 1 \times C }$.These vectors are summed to obtain the channel attention weights $M_{c}\in \bbbr^{1\times 1 \times C }$\\

\begin{align}
    Avg_{c}=MLP(F_{avg}),\\
    Max_{c}=MLP(F_{max}),\\
    M_{c}=(Max_{c}+Avg_{c}).
\end{align}

\indent Element-Wise Modulation: After obtaining the multi-scale pooling results and channel mixing outputs, we modulate the input features through element-wise multiplication. This modulation dynamically adjusts the weights of each channel based on the input feature content, thereby enhancing the model's feature representation capability.\\
\begin{equation}
    M_{D}=Conv(M_{c}).
\end{equation}
\indent Finally, a sigmoid function is applied to learn the channel attention weights, which are then multiplied with the original input features to produce the final output.\\
\begin{equation}
    M_{O}=F \times Sigmoid(M_{D}).
\end{equation}
\begin{figure}
\centering
\setlength{\abovecaptionskip}{0.cm}
\includegraphics[width=\linewidth]{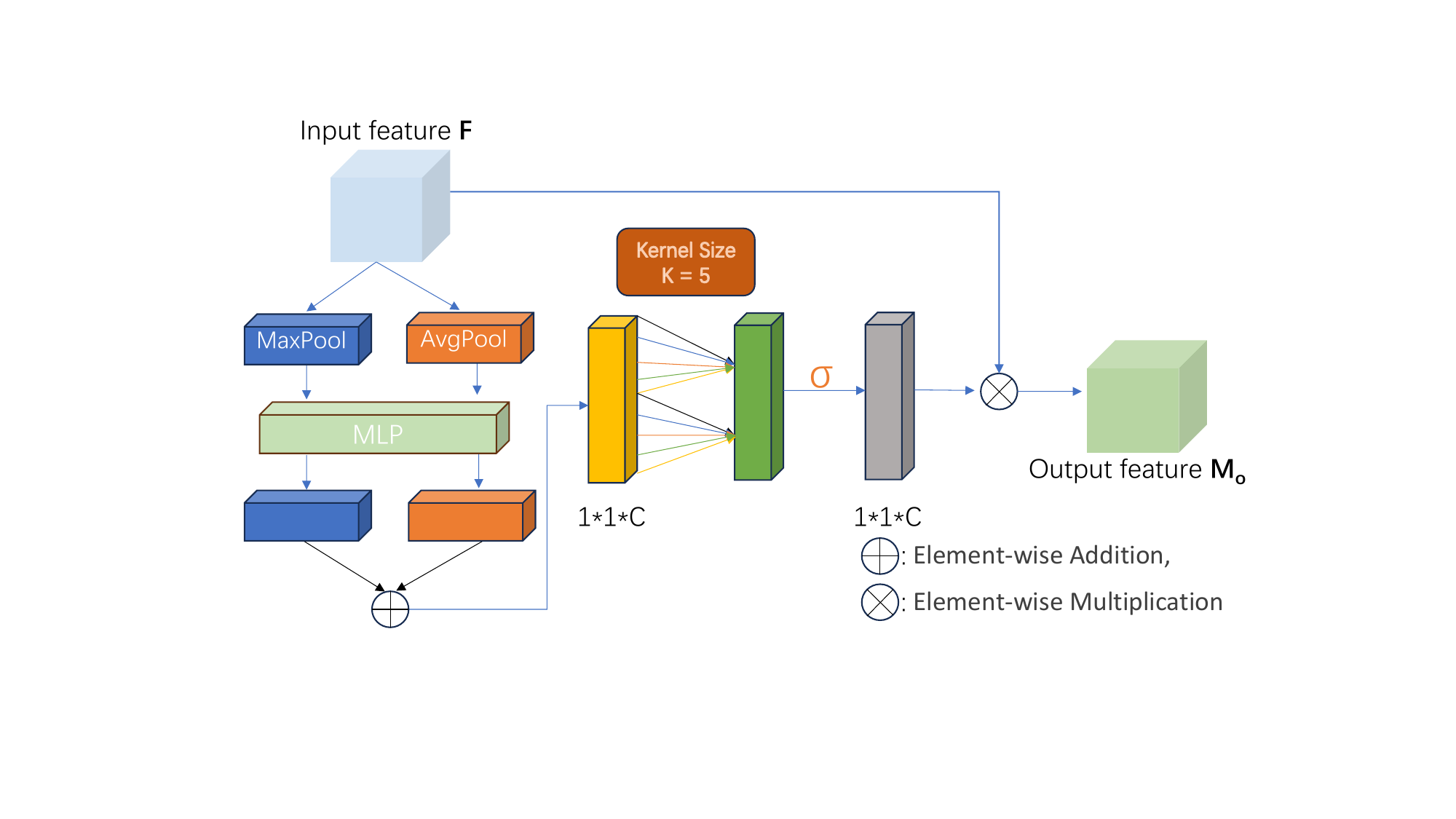}
\caption{CE attention module} \label{fig3}
\end{figure}

\subsection{Improved SAFM with Depthwise Separable Convolutions (DP-SAFM)}  
In the original SAFM module, feature modulation relies on standard convolution operations, which effectively capture spatial features but suffer from computational inefficiency and parameter redundancy. To enhance model lightweightness and reduce computational overhead, we propose replacing standard convolutions in SAFM with depthwise separable convolutions, forming the DP-SAFM module. The detailed architecture is illustrated in Figure \ref{fig2}.\\
\indent The input feature map $X\in\bbbr^{c \times H \times W }$ is evenly partitioned into four sub-features along the channel dimension:[$X_{1},X_{2},X_{3},X_{4}$] , where each $X_{i}\in \bbbr^{c/4 \times H \times W }$ retains the original spatial resolution but with a quarter of the channels.\\
\indent Multi-Scale Feature Generation:The first red part retains the original resolution $X_{1}\in \bbbr^{c/4 \times H \times W }$,the second yellow part applies 2 downsampling via max pooling, yielding $X_{2}\in \bbbr^{c/4 \times H/2 \times W/2 }$,the third Green part applies 4 downsampling, producing $X_{3}\in \bbbr^{c/4 \times H/4 \times W/4 }$,the fourth blue part applies 8 downsampling, resulting in $X_{4}\in \bbbr^{c/4 \times H/8 \times W/8 }$:\\
\begin{equation}
\begin{aligned}
X_i\in\bbbr^{C/4 \times H/2^{(i-1)} \times W/2^{(i-1)}}= MaxPool(X_i ).i>1
\end{aligned}
\end{equation}
\indent Feature Restoration via Upsampling:Each downsampled feature ($X_{2},X_{3},X_{4}$) undergoes a 3 $\times$ 3 depthwise separable convolution followed by upsampling operations to restore their spatial dimensions to $\bbbr^{c/4 \times H \times W }$.\\
\begin{equation}
\begin{aligned}
X_i \in \bbbr^{C/4 \times H \times W}= upsample(DWConv(X_i)).
\end{aligned}
\end{equation}
\indent The four multi-scale features are concatenated along the channel dimension and fused via a 1×1 convolution to integrate local and global relationships.
\begin{equation}
\begin{aligned}
\widehat{X} \in \bbbr^{C \times H \times W}= Conv(Concat(X_1,X_2,X_3,X_4) ).
\end{aligned}
\end{equation}
\indent The aggregated multi-scale features are normalized using the GELU activation function to generate an attention map. This map is then multiplied element-wise with the original input $X$ to produce the final output $Y$.
\begin{equation}
\begin{aligned}
Y \in \bbbr^{C \times H \times W}= GELU(\widehat{X}) \odot X.
\end{aligned}
\end{equation}
\indent The improved SAFM module employs depthwise convolutions for spatial feature extraction and pointwise convolutions for cross-channel information fusion. This redesign reduces computational complexity while enhancing local detail modeling, particularly suited for complex waste classification scenarios with cluttered backgrounds and diverse targets. Experiments demonstrate that the modified DP-SAFM achieves a 30\% reduction in parameters while maintaining comparable feature modulation performance to the original module, enabling efficient real-world deployment.

\subsection{Data augmentation}  
\indent Data augmentation is a critical step in deep learning model training, aiming to enhance data diversity through various transformations, thereby improving model generalization and robustness. In this study, we employ multiple augmentation techniques, including random horizontal flipping, rotation, translation, and noise injection, to improve the model’s adaptability to input variations. The original dataset’s 40 categories are each expanded by 100 images, resulting in an additional 4,000 augmented samples.\\
\indent Augmentation Strategies:\\
\indent Rotation:Training images are randomly rotated within a range of [$-90^{\circ},90^{\circ}$]to simulate viewpoint variations. This operation increases data diversity and enhances the model’s robustness to rotational changes in real-world scenarios.\\
\indent Translation:Images undergo random horizontal and vertical shifts within 10\% of their spatial dimensions. This strategy encourages the model to learn positional invariance, improving its adaptability to object location changes.\\
\indent Noise Injection:Gaussian and salt-and-pepper noise (with a standard deviation of 0.02) are added to simulate real-world sensor or environmental noise. This strengthens the model’s resilience to noisy inputs.\\
\indent In addition to the three primary data augmentation techniques mentioned above, we also implemented other augmentation strategies, such as horizontal flipping and scale-augmented rotation. Horizontal flipping increases data diversity by randomly mirroring images, while scale-augmented rotation simulates inputs at varying scales. These operations further enhance the model's adaptability to input variations, exposing it to a wider range of data transformations during training. This not only improves the model's generalization capability but also strengthens its robustness in real-world applications. Experimental results demonstrate that the application of these data augmentation strategies leads to significant improvements in both classification accuracy and robustness on the test set.\\
\begin{figure}
\centering
\setlength{\abovecaptionskip}{0.cm}
\includegraphics[width=\linewidth]{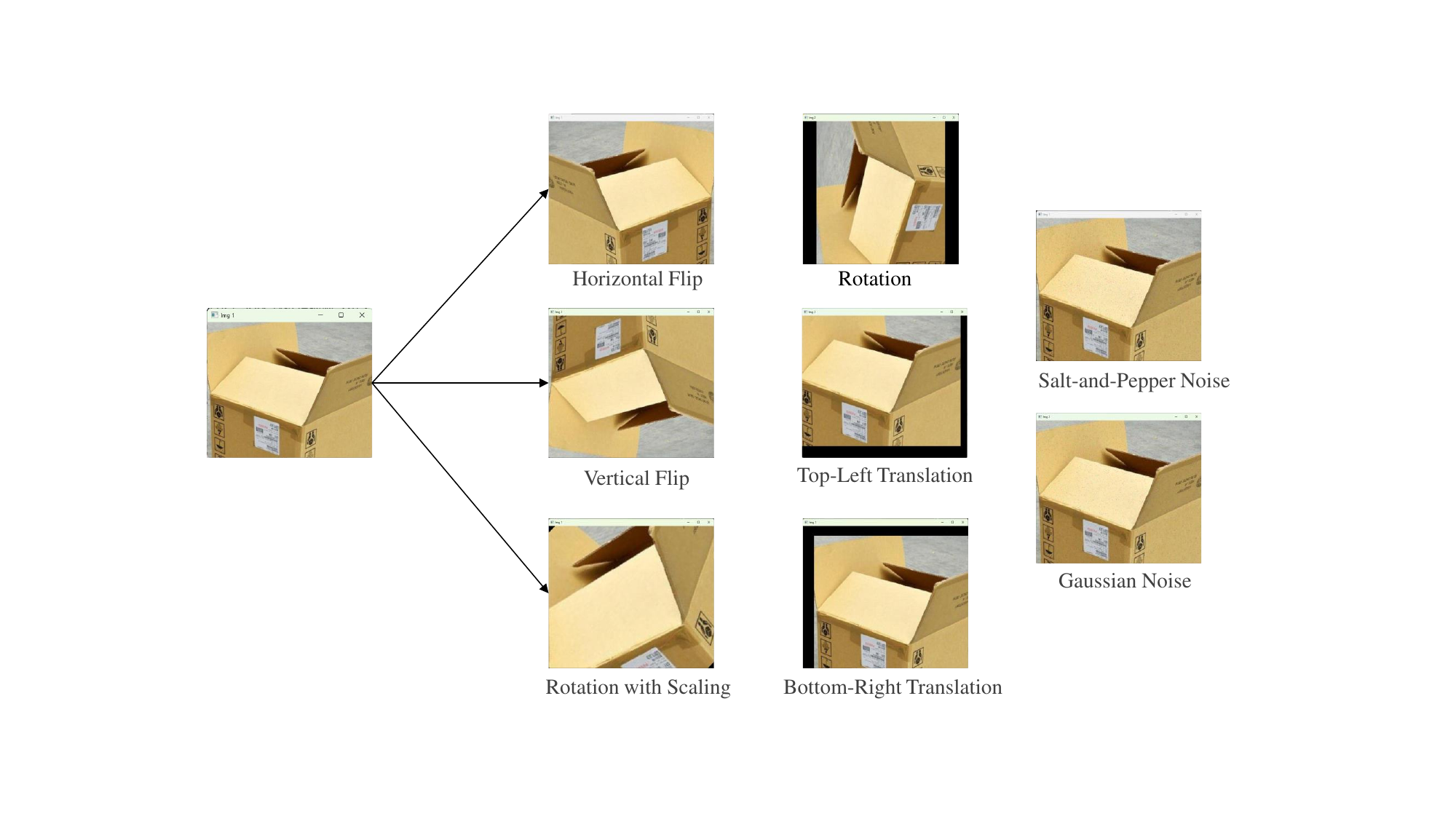}
\caption{Data Augmentation Strategies} \label{fig4}
\end{figure}

\section{Experiment and Analysis}
To validate the proposed CE-EfficientNetV2 method's superiority, it is compared with multiple state-of-the-art garbage classification approaches on two  datasets, namely, Huawei Cloud and TrashNet.
\subsection{Datasets}
\textbf{\emph{Huawei Cloud}}
\indent Released during the 2020 "Huawei Cloud AI Competition – Waste Classification Challenge," this dataset contains 14,964 images of domestic waste categorized into 4 major classes and 44 subclasses. Each subclass comprises hundreds to thousands of images with varying resolutions. The dataset follows the standard Pascal VOC format.\\
\indent \textbf{\emph{TrashNet}}
\indent TrashNet~\cite{0Classification}, developed by Gary Thung and Mindy Yang as part of Stanford University’s CS 229 Machine Learning course project, TrashNet consists of 2,527 RGB images of six waste categories: glass (501 images), paper (594), cardboard (403), plastic (482), metal (410), and general trash (137). Images were captured under controlled lighting conditions (natural or indoor) on a white poster board and resized to 512×384 pixels. The dataset aims to advance automated waste classification using machine learning, particularly CNNs, to improve recycling efficiency.
\subsection{Evaluation Metrics}  
\indent As model parameters stabilize with increasing training epochs, accuracy and loss values may still exhibit minor fluctuations. To mitigate this, the final model performance is reported as the average accuracy and loss over the last ten training epochs. The metrics are defined as:
\begin{equation}
\begin{aligned}
Accuracy_{avg}=\frac{ {\textstyle \sum_{e=40}^{49}}Accuracy_e }{10}\\
Loss_{avg}=\frac{ {\textstyle \sum_{e=40}^{49}}Loss_e }{10}
\end{aligned}
\end{equation}
where $Accuracy_e$ and $Loss_e$ denote the classification accuracy and loss on the Huawei Cloud validation set after the $e-th$ epoch.
\subsection{Implementation Details} 
\indent Model training and evaluation in this study are conducted on the Huawei Cloud waste classification dataset. The experimental platform specifications are as follows: Windows 11 operating system, NVIDIA RTX 4060 GPU, Python 3.9, and PyTorch 1.11. All experiments are trained under identical hyperparameter configurations to ensure reproducibility and consistency.\\
\indent Both datasets are split into training and test sets, with 80\% of samples per category allocated for training and 20\% for testing/validation.
\vspace{-0.8cm}
\begin{figure}
\centering
\setlength{\abovecaptionskip}{0.cm}
\includegraphics[width=\linewidth]{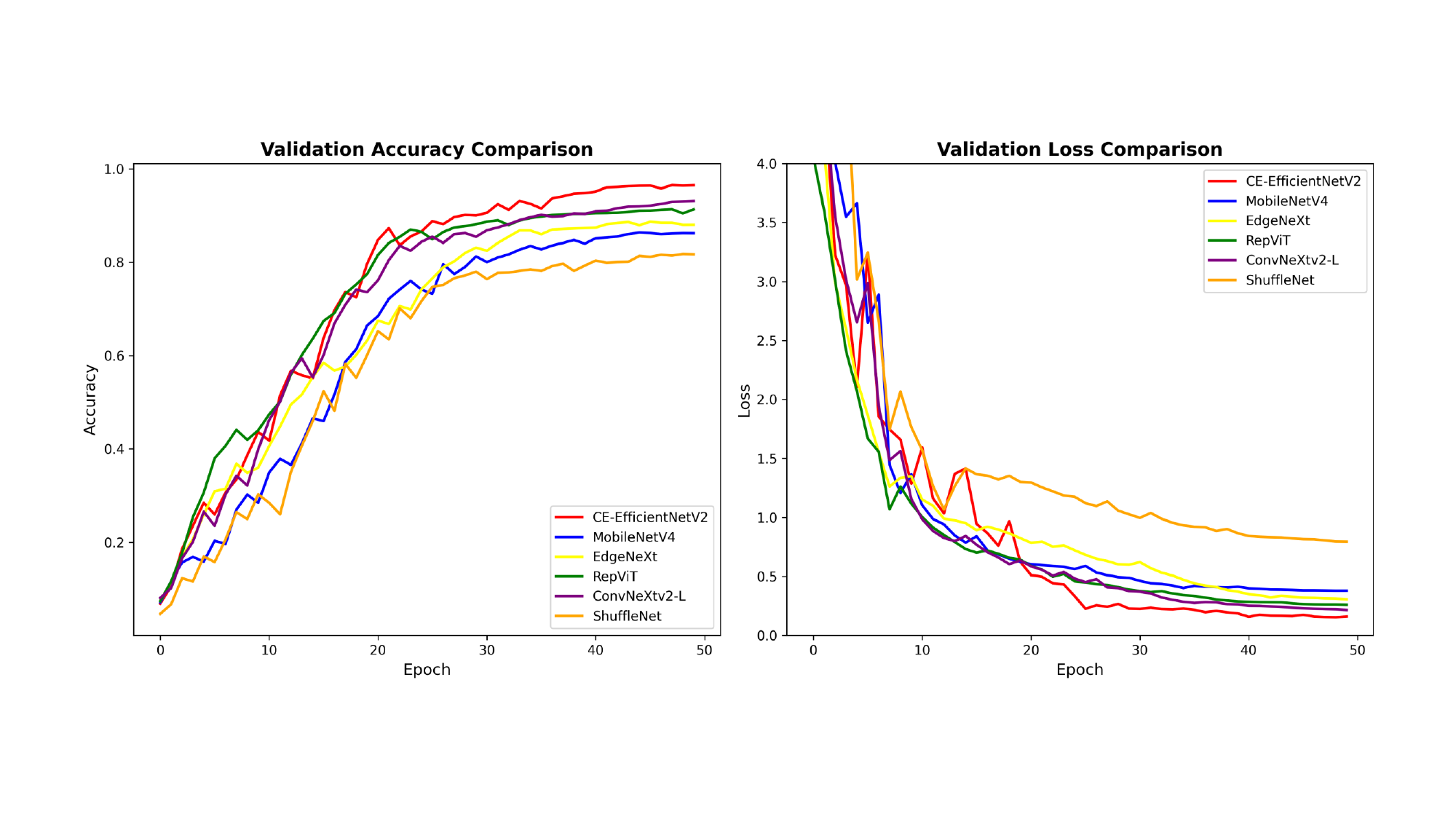}
\caption{Line Charts of Accuracy and Loss Rates for Different Models on the Huawei Cloud Dataset} \label{fig5}
\end{figure}
\vspace{-0.8cm}
\subsection{Comparison with State-of-the-art Methods} 
\indent To assess the model’s performance, we compare the proposed CE-EfficientNetV2 against several state-of-the-art neural networks, including lightweight architectures (MobileNetV4 ~\cite{2025MobileNetV4}, EdgeNeXt~\cite{maaz2022edgenext}, RepViT ~\cite{2023RepViT},ShuffleNet~\cite{zhang2018shufflenet}) and a larger-scale model (ConvNeXt V2~\cite{woo2023convnext}). All models are trained and evaluated on both the Huawei Cloud and TrashNet datasets.\\
\indent As shown in Table \ref{table1}, CE-EfficientNetV2 demonstrates excellent performance on both the Huawei Cloud and TrashNet datasets, indicating that our improvements effectively enhance the model’s capabilities. Compared to ConvNeXt V2, a similarly large-parameter model, CE-EfficientNetV2 achieves slightly higher accuracy, despite slightly inferior loss rates and recognition times. In waste classification tasks, which involve numerous categories (e.g., plastics, metals, paper) and visually similar classes (e.g., plastic bottles vs. glass bottles), high accuracy is critical to avoid misclassification and subsequent processing errors. While there is a minor gap in computational time consumption, this difference has minimal impact on overall system performance in real-world waste classification scenarios.
\begin{table}[]
\centering
\caption{Performance Comparison of Different Models on the Huawei Cloud Dataset}\label{table1}
\begin{tabular}{cccc}
\hline
\textbf{Method}   & \textbf{Accuracy\_avg} & \textbf{Loss\_avg} & \textbf{S-I Val time/ms} \\ \hline
RepViT            & 90.5\%                 & 0.274             & 16.27                    \\
ShuffleNet       & 81.7\%                 & 0.789             & \textbf{9.52}                     \\  
EdgeNeXt          & 87.6\%                 & 0.317             & 12.38                    \\
ConvNeXt V2       & 92.7\%                 & \textbf{0.227}             & 45.4                     \\
MobileNetV4-L     & 86.2\%                 & 0.386             & 10.7                     \\
CE-EfficientNetV2 & \textbf{95.4}\%                 & 0.245             & 51.3                     \\ \hline
\end{tabular}
\end{table}

\begin{table}[]
\centering
\caption{Performance Comparison of Different Models on the TrashNet Dataset}\label{table2}
\begin{tabular}{cccc}
\hline
\textbf{Method}   & \textbf{Accuracy\_avg} & \textbf{Loss\_avg} & \textbf{S-I Val time/ms} \\ \hline
RepViT            & 92.3\%                 & 0.251             & 18.3                    \\
ShuffleNet        & 83.9\%                 & 0.706             & \textbf{10.54}                   \\
EdgeNeXt          & 90.1\%                 & 0.283             & 15.1                    \\
ConvNeXt V2       & 95.2\%                 & \textbf{0.201}             & 48.2                     \\
MobileNetV4-L     & 89.4\%                 & 0.339             & 12.6                     \\
CE-EfficientNetV2 & \textbf{96.5}\%                 & 0.207             & 53.9                     \\ \hline
\end{tabular}
\end{table}
MobileNetV4,ShuffleNet and EdgeNeXt, as lightweight models, exhibit faster validation times but lower accuracy and higher loss rates, suggesting limitations in handling complex data. RepViT, a lightweight Transformer-based model, achieves decent accuracy and loss rates with low response times. However, in waste classification tasks characterized by diverse categories and cluttered backgrounds, RepViT struggles to learn sufficient discriminative features, limiting its classification precision.

Table \ref{table2} reveals that models require slightly longer inference times for TrashNet due to its larger image sizes compared to the Huawei Cloud dataset. Additionally, the smaller and simpler TrashNet dataset leads to higher average accuracy and lower average loss rates across all models. Overall, CE-EfficientNetV2 achieves robust accuracy and loss performance on both datasets, making it suitable for practical deployment in real-world applications.\\
\begin{table}[]
\centering
\caption{Accuracy Comparison of Different Improved Waste Classification Models on the TrashNet Dataset}\label{table3}
\begin{tabular}{cccc}
\hline
\textbf{Method} & \textbf{Dataset} & \textbf{Backbone} & \textbf{Accuracy} \\ \hline
Rabano~\cite{2018Common}          & TrashNet         & MobileNet         & 87.2\%            \\
MAOWL~\cite{2021Recycling}           & TrashNet         & DenseNet-121      & \textbf{99.6}\%            \\
Kennedy~\cite{kennedy2018oscarnet}         & TrashNet         & VGG-19            & 88.42\%           \\
Ma~\cite{ma2022improved}              & TrashNet         & Resnet-50         & 92.08\%           \\
Ozkaya~\cite{2019Fine}          & TrashNet         & GoogleNet         & 90.79\%           \\
Shi~\cite{shi2021waste}             & TrashNet         & VggNet            & 92.6\%            \\
Tian~\cite{tian2024garbage}            & TrashNet         & MobileNetV3       & 96.31\%            \\
Ours            & TrashNet         & CE-EfficientNetV2 & 96.5\%            \\ \hline
\end{tabular}
\end{table}

\indent Table \ref{table3} compares the performance of CE-EfficientNetV2 with other improved models on the TrashNet dataset. The benchmarked models include large-scale architectures such as DenseNet-121, VGG-19, and ResNet-50, as well as lightweight networks like GoogleNet and MobileNet. As shown in the table, the proposed CE-EfficientNetV2 achieves higher classification accuracy than all other models except DenseNet-121. However, CE-EfficientNetV2 requires far fewer parameters and computational resources than DenseNet-121, demonstrating its superior efficiency for practical deployment.
\vspace{-0.8cm}
\begin{figure}[htbp]
\centering
\setlength{\abovecaptionskip}{0.cm}
\includegraphics[width=\linewidth]{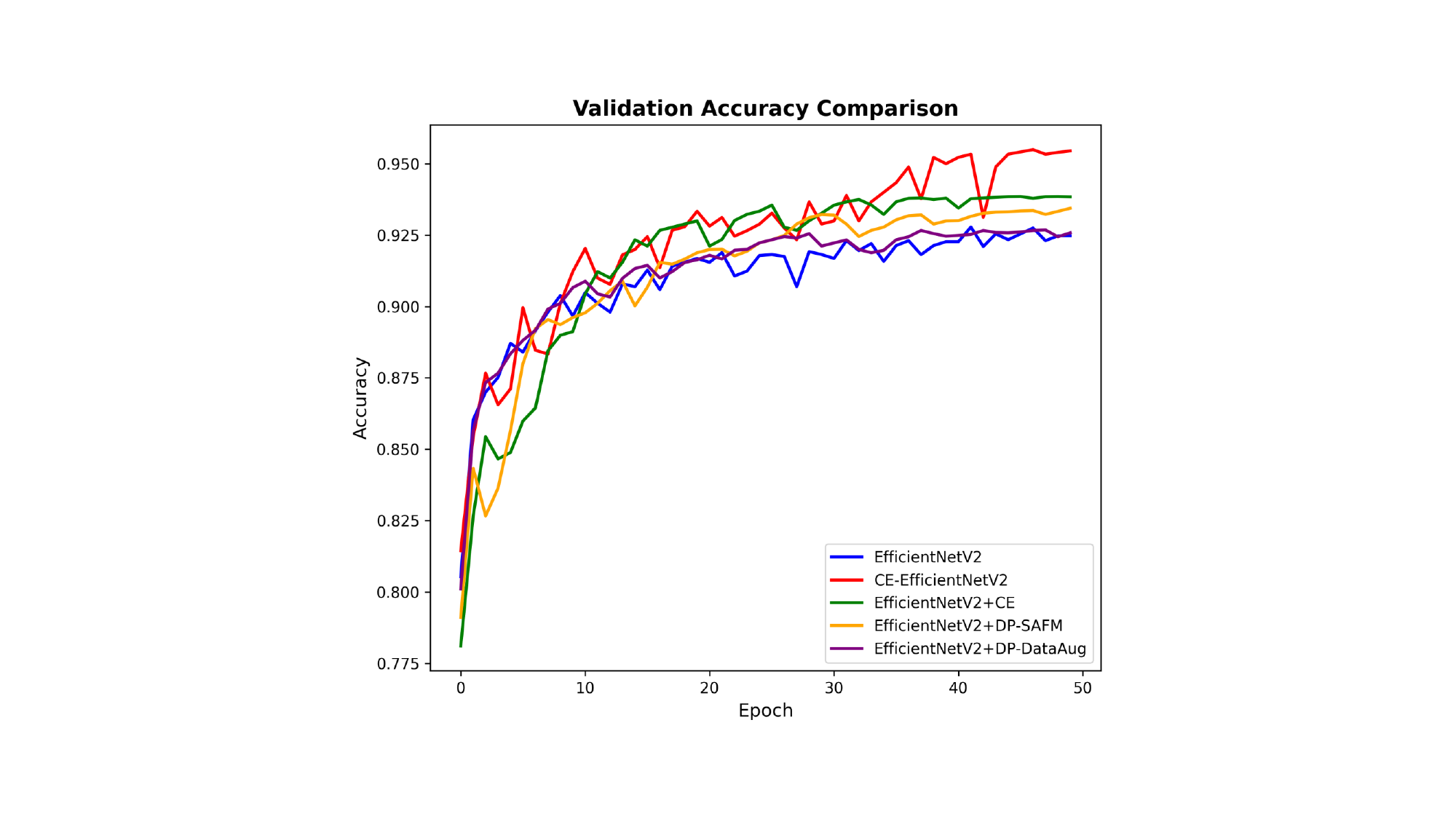}
\caption{Comparison curves for ablation experiments} \label{fig6}
\end{figure}
\vspace{-0.8cm}
\subsection{Ablation Studies and Analysis} 

\indent To validate the effectiveness of the proposed CE-EfficientNetV2 algorithm and analyze the contribution of each training strategy to classification accuracy, we conducted ablation experiments. These experiments evaluated the performance improvements brought by the CE attention mechanism, DP-SAFM module, and data augmentation strategies on the pretrained EfficientNetV2. Accuracy and model parameters were selected as evaluation metrics.\\
\indent As shown in Table 4, the CE attention mechanism improves accuracy by 1.6\% compared to the baseline, without a significant increase in model parameters or degradation in inference speed. The integration of the SAFM multi-scale feature extraction module further enhances accuracy by 1.1\%. Finally, the addition of data augmentation strategies yields an additional 0.4\% accuracy improvement. Collectively, CE-EfficientNetV2 achieves a 3.2\% increase in Top-1 accuracy over the baseline while maintaining comparable parameter counts. The ablation results demonstrate that CE-EfficientNetV2 achieves high classification accuracy with low inference latency, making it highly effective for waste classification tasks.
\begin{table}[]
\centering
\caption{Accuracy and Parameter Comparison of Different Methods for Improving EfficientNetV2-L on the Huawei Cloud Dataset}
\begin{tabular}{ccccc}
\hline
\textbf{CE} & \textbf{DP-SAFM} & \textbf{DataAug} & \textbf{Accuracy} & \textbf{Params} \\ \hline
            &                  &                  & 92.2\%            & 121M         \\
\checkmark &                  &                  & 93.8\%            & 121M           \\
            & \checkmark     &                  & 93.3\%            & 124M            \\
            &                  &  \checkmark    & 92.6\%            & 121M            \\
\checkmark & \checkmark     &  \checkmark    & 95.4\%            & 124M            \\ \hline
\end{tabular}
\end{table}

\section{Conclusion}\label{sec:con} 
\indent This paper proposes an enhanced CE-EfficientNetV2 model that integrates data augmentation techniques to improve sample diversity and generalization capabilities, introduces a novel CE attention mechanism module, and incorporates an optimized DP-SAFM module to strengthen feature extraction. These innovations collectively enhance the model’s robustness and accuracy. The proposed model demonstrates competitive performance on both the Huawei Cloud and TrashNet waste classification datasets, striking an effective balance between accuracy and computational efficiency. Experiments confirm its strong feasibility for deployment on resource-constrained devices.However, this study has limitations. It does not address multi-object recognition and classification in complex waste scenarios, and further refinements in parameter optimization remain possible. Future work will focus on developing a lightweight model capable of multi-object recognition and classification while maintaining high efficiency.

{\small
\bibliographystyle{unsrt}
\bibliography{ref}

\begin{thebibliography}{10}
\providecommand{\url}[1]{#1}
\csname url@samestyle\endcsname
\providecommand{\newblock}{\relax}
\providecommand{\bibinfo}[2]{#2}
\providecommand{\BIBentrySTDinterwordspacing}{\spaceskip=0pt\relax}
\providecommand{\BIBentryALTinterwordstretchfactor}{4}
\providecommand{\BIBentryALTinterwordspacing}{\spaceskip=\fontdimen2\font plus
\BIBentryALTinterwordstretchfactor\fontdimen3\font minus \fontdimen4\font\relax}
\providecommand{\BIBforeignlanguage}[2]{{%
\expandafter\ifx\csname l@#1\endcsname\relax
\typeout{** WARNING: IEEEtran.bst: No hyphenation pattern has been}%
\typeout{** loaded for the language `#1'. Using the pattern for}%
\typeout{** the default language instead.}%
\else
\language=\csname l@#1\endcsname
\fi
#2}}
\providecommand{\BIBdecl}{\relax}
\BIBdecl

\bibitem{2024A}
Y.~Ma and R.~Cheng, ``A discussion on the environmental issues and management approaches of municipal solid waste landfills,'' \emph{Environmental Bulletin of Heilongjiang}, vol.~37, no.~1, pp. 145--147, 2024.

\bibitem{2023A}
C.~Yu, Y.~Rao, Q.~Liao, Q.~Lan, and Y.~Wang, ``A study on the effectiveness of co-processing aged landfill waste in waste-to-energy plants: A case study of a waste incineration power plant,'' \emph{Environmental Journal of Sichuan}, vol.~42, no.~5, pp. 15--19, 2023.

\bibitem{2019Fully}
R.~Xu, ``Fully intelligent trash bin design using multi-drive control,'' \emph{Electric Tools}, no.~3, p.~3, 2019.

\bibitem{2004Distinctive}
D.~G. Lowe, ``Distinctive image features from scale-invariant keypoints,'' \emph{International Journal of Computer Vision}, vol.~60, no.~2, pp. 91--110, 2004.

\bibitem{0Body}
E.~Corvee and F.~Bremond, ``Body parts detection for people tracking using trees of histogram of oriented gradient descriptors,'' in \emph{2010 7th IEEE International Conference on Advanced Video and Signal Based Surveillance}.

\bibitem{1995Support}
C.~Cortes and V.~Vapnik, ``Support-vector networks,'' \emph{Machine Learning}, vol.~20, no.~3, pp. 273--297, 1995.

\bibitem{gao2024exploring}
B.~Gao, J.~Ren, F.~Shen, M.~Wei, and Z.~Huang, ``Exploring warping-guided features via adaptive latent diffusion model for virtual try-on,'' in \emph{2024 IEEE International Conference on Multimedia and Expo (ICME)}.\hskip 1em plus 0.5em minus 0.4em\relax IEEE, 2024, pp. 1--6.

\bibitem{2001A}
H.~B. Mitchell and P.~A. Schaefer, ``A \textquotedblleft soft\textquotedblright{} k‐nearest neighbor voting scheme,'' \emph{International Journal of Intelligent Systems}, vol.~16, no.~4, pp. 459--468, 2001.

\bibitem{2016Deep}
K.~He, X.~Zhang, S.~Ren, and J.~Sun, ``Deep residual learning for image recognition,'' \emph{IEEE}, 2016.

\bibitem{weng2024enhancing}
W.~Weng, M.~Wei, J.~Ren, and F.~Shen, ``Enhancing aerial object detection with selective frequency interaction network,'' \emph{IEEE Transactions on Artificial Intelligence}, vol.~1, no.~01, pp. 1--12, 2024.

\bibitem{li2024lr}
H.~Li, R.~Zhang, Y.~Pan, J.~Ren, and F.~Shen, ``Lr-fpn: Enhancing remote sensing object detection with location refined feature pyramid network,'' \emph{arXiv preprint arXiv:2404.01614}, 2024.

\bibitem{2021Research}
G.~Yaqing and B.~Ge, ``Research on lightweight convolutional neural network in garbage classification,'' \emph{IOP Conference Series Earth and Environmental Science}, vol. 781, no.~3, p. 032011, 2021.

\bibitem{2017MobileNets}
A.~G. Howard, M.~Zhu, B.~Chen, D.~Kalenichenko, W.~Wang, T.~Weyand, M.~Andreetto, and H.~Adam, ``Mobilenets: Efficient convolutional neural networks for mobile vision applications,'' 2017.

\bibitem{2019EfficientNet}
M.~Tan and Q.~V. Le, ``Efficientnet: Rethinking model scaling for convolutional neural networks,'' 2019.

\bibitem{2021EfficientNetV2}
------, ``Efficientnetv2: Smaller models and faster training,'' 2021.

\bibitem{2020ECA}
Q.~Wang, B.~Wu, P.~Zhu, P.~Li, and Q.~Hu, ``Eca-net: Efficient channel attention for deep convolutional neural networks,'' in \emph{2020 IEEE/CVF Conference on Computer Vision and Pattern Recognition (CVPR)}, 2020.

\bibitem{shen2023pbsl}
F.~Shen, X.~Shu, X.~Du, and J.~Tang, ``Pedestrian-specific bipartite-aware similarity learning for text-based person retrieval,'' in \emph{Proceedings of the 31th ACM International Conference on Multimedia}, 2023.

\bibitem{sun2023spatially}
L.~Sun, J.~Dong, J.~Tang, and J.~Pan, ``Spatially-adaptive feature modulation for efficient image super-resolution,'' in \emph{Proceedings of the IEEE/CVF international conference on computer vision}, 2023, pp. 13\,190--13\,199.

\bibitem{0Classification}
M.~Yang and G.~Thung, ``Classification of trash for recyclability status.''

\bibitem{2025MobileNetV4}
D.~Qin, C.~Leichner, M.~Delakis, M.~Fornoni, S.~Luo, F.~Yang, W.~Wang, C.~Banbury, C.~Ye, and B.~Akin, ``Mobilenetv4: Universal models forthemobile ecosystem,'' in \emph{European Conference on Computer Vision}, 2025.

\bibitem{maaz2022edgenext}
M.~Maaz, A.~Shaker, H.~Cholakkal, S.~Khan, S.~W. Zamir, R.~M. Anwer, and F.~Shahbaz~Khan, ``Edgenext: efficiently amalgamated cnn-transformer architecture for mobile vision applications,'' in \emph{European conference on computer vision}.\hskip 1em plus 0.5em minus 0.4em\relax Springer, 2022, pp. 3--20.

\bibitem{2023RepViT}
A.~Wang, H.~Chen, Z.~Lin, J.~Han, and G.~Ding, ``Repvit: Revisiting mobile cnn from vit perspective,'' \emph{IEEE}, 2023.

\bibitem{shen2025long}
F.~Shen, C.~Wang, J.~Gao, Q.~Guo, J.~Dang, J.~Tang, and T.-S. Chua, ``Long-term talkingface generation via motion-prior conditional diffusion model,'' \emph{arXiv preprint arXiv:2502.09533}, 2025.

\bibitem{shen2024imagpose}
F.~Shen and J.~Tang, ``Imagpose: A unified conditional framework for pose-guided person generation,'' in \emph{The Thirty-eighth Annual Conference on Neural Information Processing Systems}, 2024.

\bibitem{zhang2018shufflenet}
X.~Zhang, X.~Zhou, M.~Lin, and J.~Sun, ``Shufflenet: An extremely efficient convolutional neural network for mobile devices,'' in \emph{Proceedings of the IEEE conference on computer vision and pattern recognition}, 2018, pp. 6848--6856.

\bibitem{woo2023convnext}
S.~Woo, S.~Debnath, R.~Hu, X.~Chen, Z.~Liu, I.~S. Kweon, and S.~Xie, ``Convnext v2: Co-designing and scaling convnets with masked autoencoders,'' in \emph{Proceedings of the IEEE/CVF conference on computer vision and pattern recognition}, 2023, pp. 16\,133--16\,142.

\bibitem{2018Common}
S.~L. Rabano, M.~K. Cabatuan, E.~Sybingco, E.~P. Dadios, and E.~J. Calilung, ``Common garbage classification using mobilenet,'' in \emph{2018 IEEE 10th International Conference on Humanoid, Nanotechnology, Information Technology,Communication and Control, Environment and Management (HNICEM)}, 2018.

\bibitem{2021Recycling}
W.~L. Mao, W.~C. Chen, C.~T. Wang, and Y.~H. Lin, ``Recycling waste classification using optimized convolutional neural network,'' \emph{Resources, Conservation and Recycling}, vol. 164, no. 000, 2021.

\bibitem{kennedy2018oscarnet}
T.~Kennedy, ``Oscarnet: Using transfer learning to classify disposable waste,'' \emph{CS230 Report: Deep Learning. Stanford University, CA, Winter}, 2018.

\bibitem{shen2024imagdressing}
F.~Shen, X.~Jiang, X.~He, H.~Ye, C.~Wang, X.~Du, Z.~Li, and J.~Tang, ``Imagdressing-v1: Customizable virtual dressing,'' \emph{arXiv preprint arXiv:2407.12705}, 2024.

\bibitem{shen2023advancing}
F.~Shen, H.~Ye, J.~Zhang, C.~Wang, X.~Han, and W.~Yang, ``Advancing pose-guided image synthesis with progressive conditional diffusion models,'' \emph{arXiv preprint arXiv:2310.06313}, 2023.

\bibitem{shen2024boosting}
F.~Shen, H.~Ye, S.~Liu, J.~Zhang, C.~Wang, X.~Han, and W.~Yang, ``Boosting consistency in story visualization with rich-contextual conditional diffusion models,'' \emph{arXiv preprint arXiv:2407.02482}, 2024.

\bibitem{ma2022improved}
X.~Ma, Z.~Li, and L.~Zhang, ``An improved resnet-50 for garbage image classification,'' \emph{Tehni{\v{c}}ki vjesnik}, vol.~29, no.~5, pp. 1552--1559, 2022.

\bibitem{2019Fine}
U.~Ozkaya and L.~Seyfi, ``Fine-tuning models comparisons on garbage classification for recyclability,'' 2019.

\bibitem{shi2021waste}
C.~Shi, C.~Tan, T.~Wang, and L.~Wang, ``A waste classification method based on a multilayer hybrid convolution neural network,'' \emph{Applied Sciences}, vol.~11, no.~18, p. 8572, 2021.

\bibitem{tian2024garbage}
X.~Tian, L.~Shi, Y.~Luo, and X.~Zhang, ``Garbage classification algorithm based on improved mobilenetv3,'' \emph{IEEE Access}, 2024.

\end{thebibliography}
}
\end{document}